\documentclass[10pt,twocolumn,letterpaper]{article}

\usepackage{cvpr}              % To produce the CAMERA-READY version
\usepackage{graphicx}
\usepackage{amsmath}
\usepackage{amssymb}
\usepackage{booktabs}
\usepackage{amsfonts}       % blackboard math symbols
\usepackage{nicefrac}       % compact symbols for 1/2, etc.
\usepackage{microtype}      % microtypography
\usepackage{xcolor}         % colors
\usepackage{graphicx}
\usepackage{diagbox}
\usepackage{multicol}
\usepackage{enumerate}
\usepackage{times}
\usepackage{epsfig}
\usepackage{graphicx}
\usepackage{amsmath}
\usepackage{amssymb}
\usepackage{threeparttable}
\usepackage{enumitem}
\usepackage{multirow}
\usepackage{color}
\usepackage{array}

\usepackage{ulem}
\newcommand{\li}{\uline{\hspace{0.5em}}}
\usepackage[colorlinks,linkcolor=red]{hyperref}
\usepackage[numbers,sort&compress]{natbib}
\setenumerate[1]{leftmargin = 10pt, itemsep=0pt,partopsep=0pt,parsep=-0pt,topsep=-0pt}
\usepackage{hyperref}

\newcommand{\PreserveBackslash}[1]{\let\temp=\\#1\let\\=\temp}

\usepackage[capitalize]{cleveref}
\crefname{section}{Sec.}{Secs.}
\Crefname{section}{Section}{Sections}
\Crefname{table}{Table}{Tables}
\crefname{table}{Tab.}{Tabs.}

\def\cvprPaperID{5008} % *** Enter the CVPR Paper ID here
\def\confName{CVPR}
\def\confYear{2022}
\def\hlinew#1{%
 \noalign{\ifnum0=`}\fi\hrule \@height #1 \futurelet
 \reserved@a\@xhline}
 
\begin{document}

\title{A Two-Stage Real Image Deraining Method for GT-RAIN Challenge\\ CVPR 2023 Workshop UG$^{\textbf{2}}$+ Track 3}

\author{Yun Guo, Xueyao Xiao, Xiaoxiong Wang, Yi Li, Yi Chang\footnotemark[1],  Luxin Yan\footnotemark[1]\\
Huazhong University of Science and Technology\\
{\tt\small \{guoyun, xiaoxueyao, wangxiaoxiong, li\li yi, yichang, yanluxin\}@hust.edu.cn}}
\maketitle

%%%%%%%%% ABSTRACT
\begin{abstract}
   In this technical report, we briefly introduce the solution
of our team HUST\li VIE for GT-Rain Challenge in CVPR 2023 UG$^{2}$+ Track 3. In this task,
we propose an efficient two-stage framework to reconstruct
a clear image from rainy frames. Firstly, a low-rank based video deraining method is utilized to generate pseudo GT, which fully takes the advantage of multi and aligned rainy frames. Secondly, a transformer-based single image deraining network Uformer is implemented to pre-train on large real rain dataset and then fine-tuned on pseudo GT to further improve image restoration. Moreover, in terms of visual pleasing effect, a comprehensive image processor module is utilized at the end of pipeline. Our overall framework is elaborately designed and able to handle both heavy rainy and foggy sequences provided in the final testing phase. Finally, we rank \textbf{1st on the average structural similarity (SSIM) and rank 2nd on the average peak signal-to-noise ratio (PSNR)}. Our code is available at \url{{https://github.com/yunguo224/UG2_Deraining}}.
\end{abstract}

%%%%%%%%% BODY TEXT
\vspace{-10pt}
\section{Introduction}
\begin{figure}[t]
 \vspace{0cm}  %调整图片与上文的垂直距离
\setlength{\abovecaptionskip}{0 cm}   %调整图片标题与图距离
\setlength{\belowcaptionskip}{-0.4 cm}   %调整图片标题与下文距离
  \centering
     \includegraphics[width=1.00\linewidth]{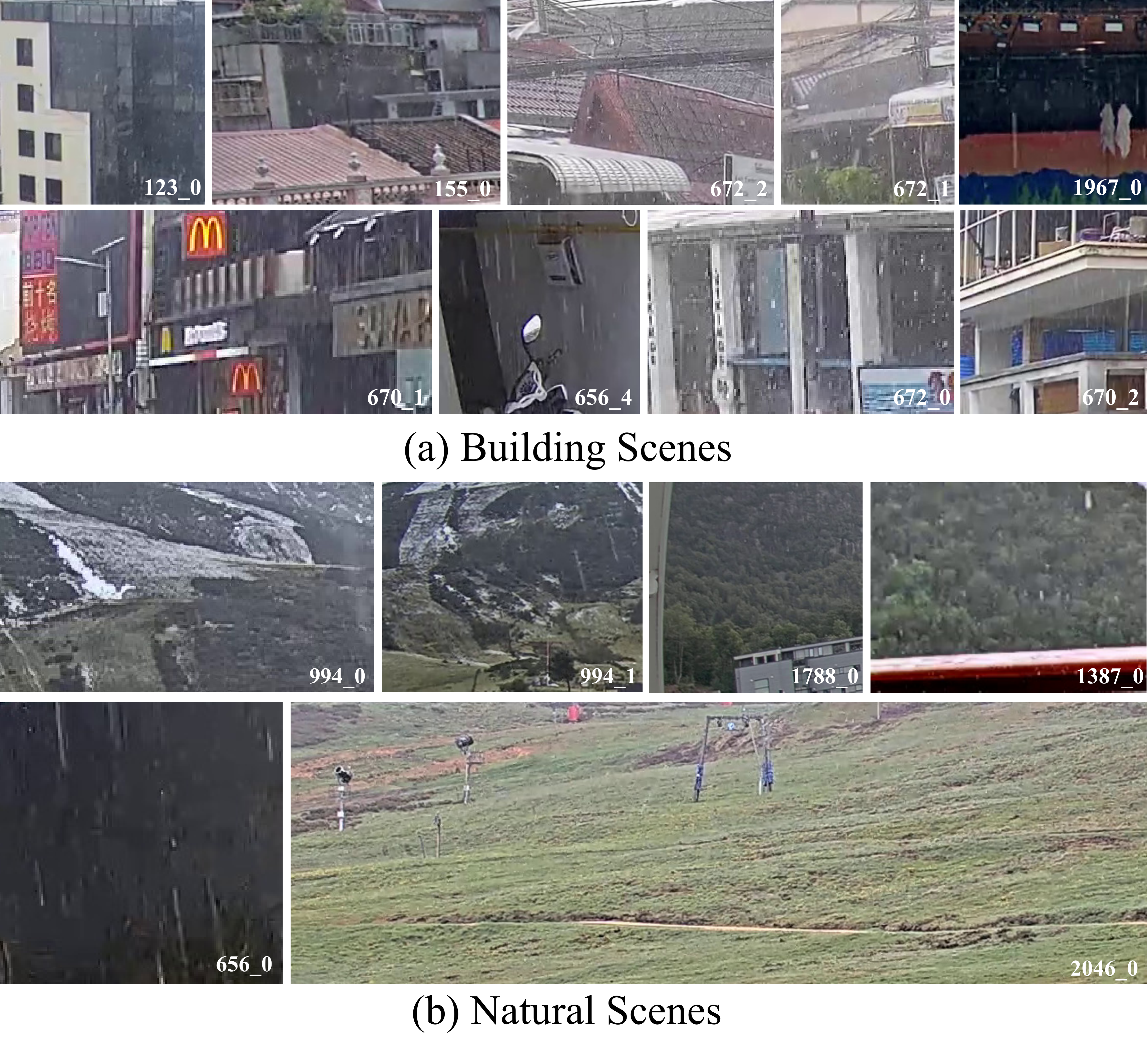}
  \caption{Overview of 15 Rainy sequences provided in testing phase. We split them into two categories by image scene, (a) Building scenes and (b) Natural scenes.}
  \label{Overview}
\end{figure}

This technical report clarifies our solutions to CVPR 2023 UG$^{2}$+ Track 3 GT-Rain Challenge, which aims to remove degradations induced by rain from images. In this task, 15 different degraded rainy sequences of all kinds of scenes in the real world, as shown in Fig. \ref{Overview}, are provided in the final testing phase and participants should design and implement rain removal methods to obtain better restoration results, which is quantitatively evaluated by PSNR and SSIM. 

We design a two-stage real image deraining method to handle the provided degraded sequences. In the first stage, we utilize a low-rank based video deraining method, which can generate higher-quality pseudo GT with better rain removal and image structure preserving. In the second stage, we implement a transformer-based single image deraining network Uformer to pre-train on a large paired real rain dataset, and then fine-tuned on pseudo GT to further improve image restoration.

Finally, our approach shows mighty competitive ability on restoring on GT-Rain testing sequences in the final phase, which ranks first in terms of SSIM and second in terms of PSNR on the final leaderboard. 

The technical report includes the following sections:
\begin{itemize}
\item Detailed analysis of datasets (Section \ref{sec2}). In this section, we conduct deep analysis on the properties of provided testing sequences, such as background alignment, veiling effect and histogram of pixel value distribution.
\item Overview of our proposed deraining method (Section \ref{sec3}). We illustrate the overall pipeline of our two-stage deraining method and details of important module.
 \item Experiment and results(Section \ref{sec4}). Extensive experiments verify the superiority of the proposed dataset and deraining method over state-of-the-art.
\end{itemize}

\begin{figure*}[htbp]
  \centering
     \includegraphics[width=1.00\linewidth]{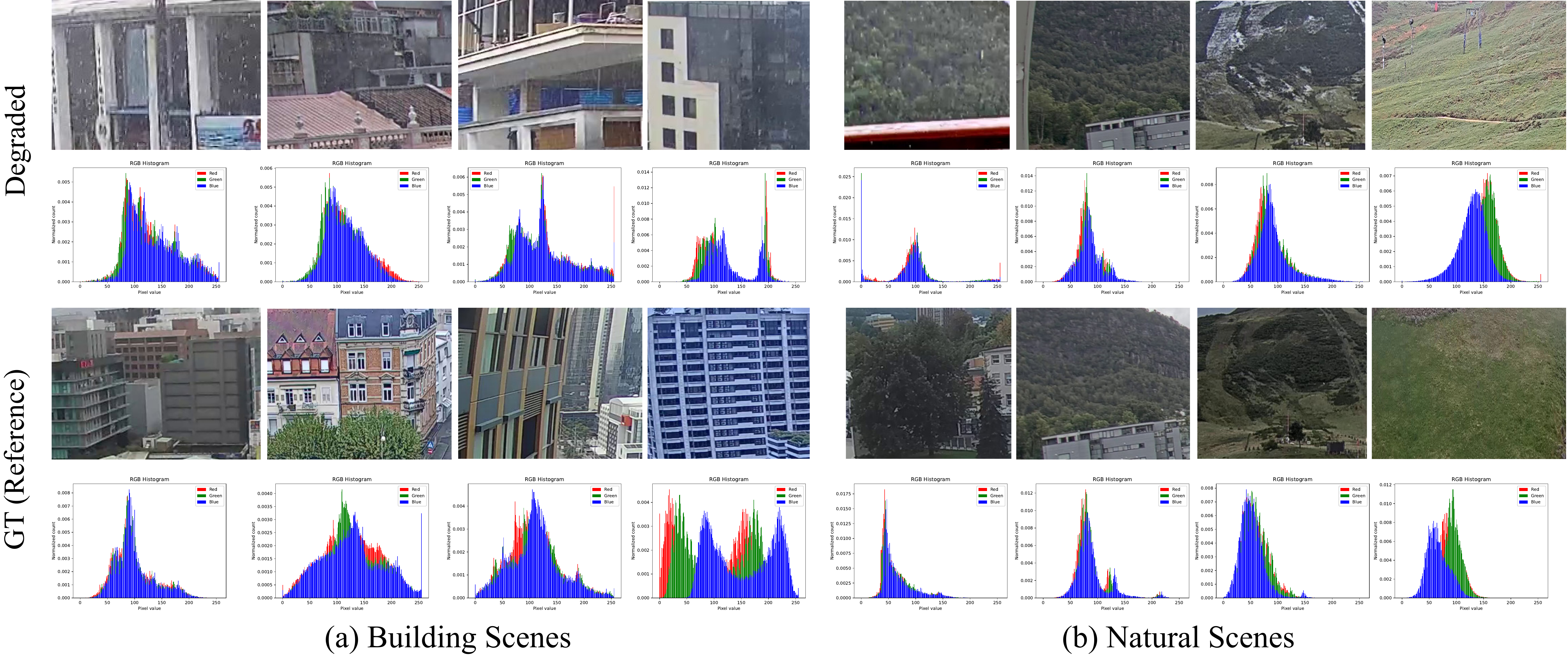}
  \caption{Histograms of the degraded samples from final testing sequences and reference GT from WeatherStream dataset. The reference GT has similar background with degraded cases, helps to improve  restoration in statistic level. (a) Building scenes have more structural elements and wider distribution of pixel value. (b) Natural scenes have more textured elements and more concentrated distribution of pixel value. }
  \label{Analysis}
\end{figure*}

\section{Dataset Analysis}\label{sec2}
In this section, we conduct detailed analysis on the provided 15 testing sequences and GT-Rain dataset \cite{ba2022gt-rain}, and several important properties of this datasets have been discovered, such as background alignment, veiling effect and distribution of pixel value, which strongly inspires us to propose the two-stage image deraining solution for this task.    
\subsection{Multi-Frame Alignment}\label{sec2.1}
According to the testing sequences provided in the final phase, each sequence contains 300 rainy frames with the same scene, which satisfies strict background alignment. As claimed in \cite{ba2022gt-rain}, background unchanging portions are cropped and camera/local motions are carefully realigned, which means that the dynamic rain streaks of each frame could be easily efficiently removed by video-based deraining method. Taking advantage of multi-frame information, pseudo GT could be generated to help single image deraining.  

\subsection{Statistical Histogram}\label{sec2.3}
By splitting 15 testing sequences into building and natural scenes, we further explore the distribution of pixel value for different scenes. As illustrated in Fig. \ref{Analysis}, we plot the histograms of typical sequences for comparison in statistical level. Specifically, the degraded cases and their corresponding RGB histograms from building and natural scenes are on the top, which have different distribution of pixel value. Because of the absence of corresponding GT, we select reference GT which shares very similar background in GT-Rain and WeatherStream-Rain datasets on the bottom. Horizontally, the cases from building scenes have more structural elements, such as straight edges, whose histograms share wider distribution of pixel value. However, other cases from natural scenes have more complex textured elements like leaves, with more concentrated distribution of pixel value.

\subsection{Veiling Effect Phenomenon}\label{sec2.2}
Except for common rain streak, the veiling effect is a special phenomenon in GT-Rain dataset \cite{ba2022gt-rain}. The appearance of veiling effect contains two different degradations simultaneously, rain and haze. It occurs significantly in several cases of testing sequences, such as 672\li 0, 672\li 1 and 672\li 2, as shown in Fig. \ref{Overview}. Degraded by veiling effect, the original image suffers from low contrast, unclear edge and stochastic blocking of rain streaks at the same time, which is a tough task to handle because of the entanglement of multi degradations. As has been pointed out, popular streak-only rain dataset \cite{yang2017deep, fu2017removing, wang2019spatial} could not deal with such veiling effect, leading to the remaining haze effect. Generally speaking, except for GT-Rain, we should search for another paired real rain dataset with veiling effect to improve the performance on the final 15 testing sequences. 

Recently, we notice that GT-Rain team publishes their latest work WeatherStream \cite{zhang2023weatherstream}, which is a larger extension of their previous work GT-Rain, including two typical adverse weather rain and snow. The training set of WeatherStream-Rain has 473 sequences with more than 142K frames, which is much larger than GT-Rain (89 sequences, 26.7K frames). Fortunately, the data source and collection strategy of GT of WeatherStream is much similar to that of GT-Rain, which contains abundant cases with significant veiling effect. As a result, we mix the both GT-Rain and WeatherStream-Rain as a full large dataset for pre-training, which could provide sufficient knowledge on image deraining, guaranteeing the basic performance on testing sequences. 
\begin{figure*}[t]
  \centering
     \includegraphics[width=1.00\linewidth]{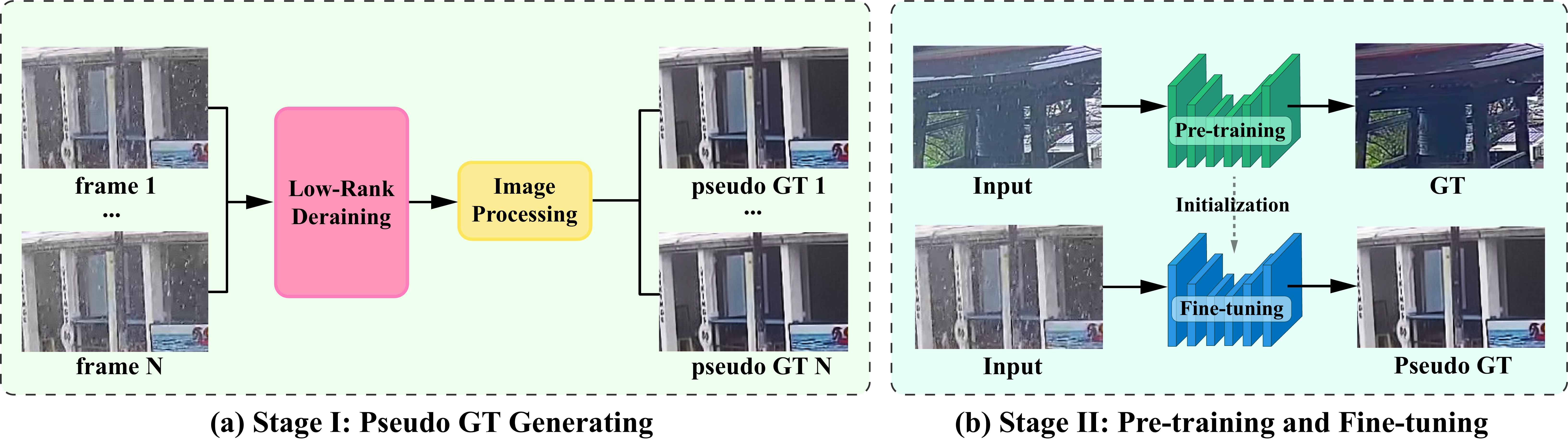}
  \caption{Framework of our proposed two-stage deraining method. (a) A low-rank based video deraining method and image processing technique are utilized to generate pseudo GT. (b) A transformer-based single image deraining network Uformer is implemented to pre-train on large real rain datasets and then fine-tuned on pseudo GT. }
  \label{Framework}
\end{figure*}

Histogram reveals the basic characteristic of pixel value distribution, which helps to reasonably minimize the distance between our final restoration results and corresponding GT not provided. Note that the GT selected from other datasets could only perform as reference, rather than the target distribution. But they also play an important role because the provided 15 sequences and reference GT share the same data source, collection strategy and similar background, especially for several typical cases in natural scenes (collected at the same position but different time and view). 

\section{Overview of Methods}\label{sec3}
Our overall framework contains two important stages. In stage I, we utilize a low-rank based video deraining method and image processing technique to predict pseudo GT. In stage II, a transformer-based single image deraining network Uformer is implemented to pre-train on GT-Rain and WeatherStream-Rain datasets and then fine-tuned on pseudo GT to further improve image restoration.

\subsection{Pseudo GT Generation}\label{sec3.1}
With analysis in Sec \ref{sec2.1}, pseudo GT could be generated by multi frames to help single image deraining. We utilize a low-rank based video deraining method LLRT \cite{chang2017hyper} to generate high-quality pseudo GT and pipeline is illustrated as stage I in Fig. \ref{Framework}(a). Given the rainy sequence, the key is how to properly obtain the paired GT. The existing methods simply employ the naive filtering technique benefiting from the temporal consistency of the static background, but leading to over-smooth issue. Low-rank based video deraining method could achieve better rain removing and image structure preserving than such filtering technique.

\subsection{Image Processing Module}\label{sec3.3}
The Low-rank based video deraining has achieved good results, but based on the predicted outcome of model, more effort could be made for improve the image details and visibility. As clarified in Sec \ref{sec2.3}, histograms of reference GT play an important role in improving final performance. By selecting reference GT sharing similar background with 15 testing sequences, the differences still exist in terms of pixel value distribution. In order to minimize the difference between predicted outcome and final result, we add an image processing module to moderate the output of each sequence individually according to the histogram of reference GT. 

This module contains two basic methods, contrast enhancement and sharpen technique, which are commonly used in image processing to adjust image details and visibility. In terms of contrast enhancement, we utilize Gamma Correction by following Eq. (\ref{Contrast}):
\begin{equation}
\setlength{\abovedisplayskip}{2pt}
\setlength{\belowdisplayskip}{2pt}
{I_{enhance}}= {(I/255)}^{\gamma}*255.
\label{Contrast}
 \end{equation}
where $I$ is original result, $I_{enhance}$ means the enhancing result and $\gamma$ is the exponent to adjust enhancement level.  

In terms of Sharpen, we utilize Unsharpen Mask method by following Eq. (\ref{Sharpen}):
\begin{equation}
\setlength{\abovedisplayskip}{2pt}
\setlength{\belowdisplayskip}{2pt}
{I_{sharpen}}= I-w*{G(I, k)}.
\label{Sharpen}
 \end{equation}
where $I_{sharpen}$ means the sharpen result, $w$ is the weight (set as  0.5), $G(\cdot)$ is Gaussian blur function and $k$ is the Gaussian kernel size to control sharpening level. 

\subsection{Pre-training and Fine-tuning}\label{sec3.2}
As mentioned in Sec \ref{sec2.2}, in order to handle both rain streak and veiling effect, we utilize the strategy of pre-training and fine-tune. The process of stage II is described in Fig. \ref{Framework}(b). Specifically, we first pre-train an experienced model on a large real rain dataset, which has abundant knowledge on restoring tough rain degradations under various backgrounds. Meanwhile, to improve the performance on final testing sequences, the pre-trained model will be fine-tuned on degraded frames and pseudo GT pairs, which is reasonable because pseudo GT offers a clear direction for abundant pre-trained knowledge to transfer instead of blindly depending on the generalization of model. Moreover, considering both representation ability and efficiency, we select Uformer \cite{wang2022uformer}, a general U-shaped transformer for image restoration as backbone network, which has excellent performance on image deraining tasks. More details are discussed in Sec \ref{sec3.4}. 
 
 \begin{figure*}[t]
  \vspace{0cm}  %调整图片与上文的垂直距离
\setlength{\belowcaptionskip}{-0.4 cm}   %调整图片标题与下文距离
  \centering
     \includegraphics[width=1.00\linewidth]{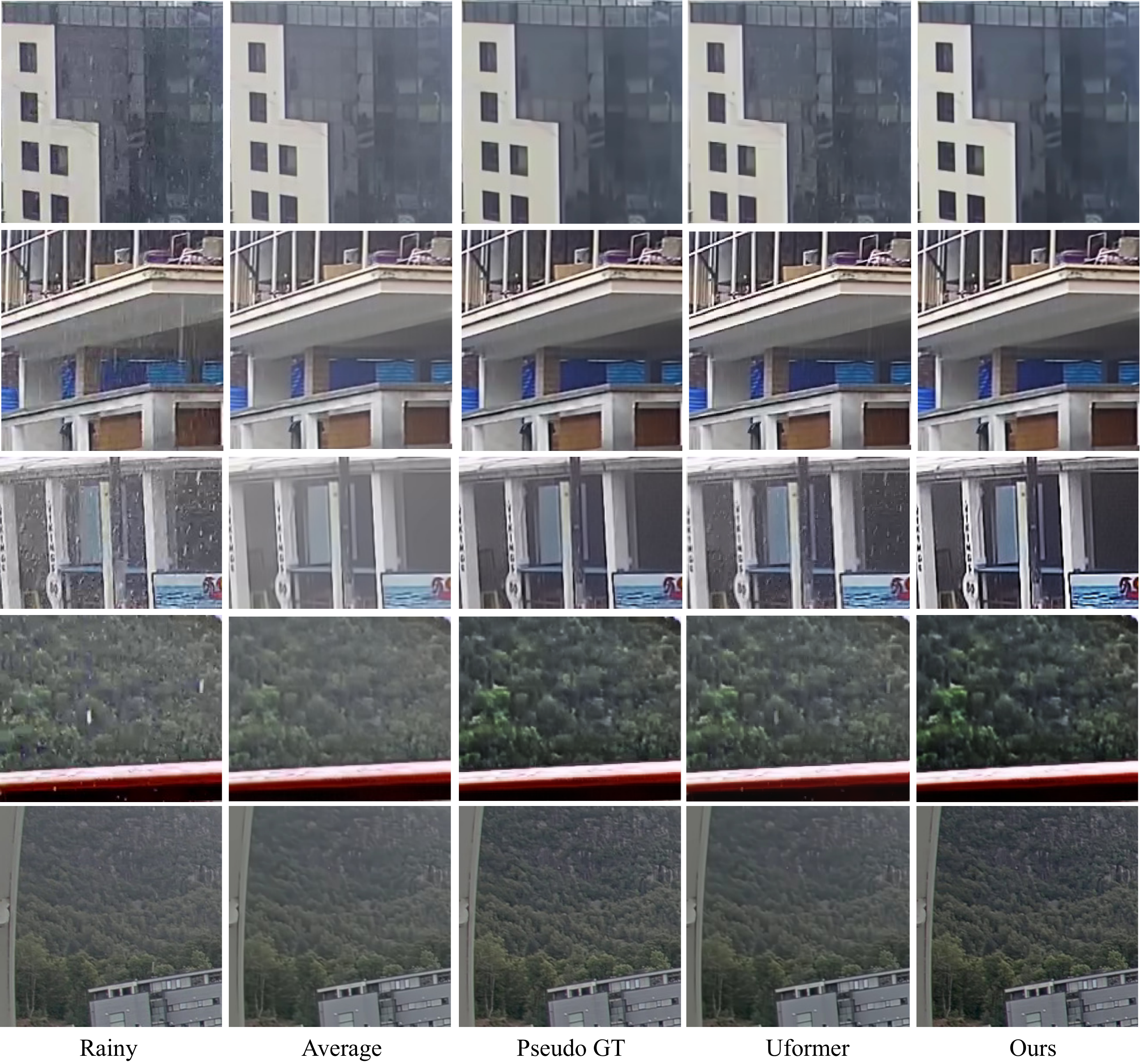}
  \caption{Deraining results of testing sequences. Our final results perform better at preserving image structure and removing both streaks and haze-like veiling effect.}
  \label{Visualization}
\end{figure*}

\subsection{Implementation Details}\label{sec3.4}
The framework is implemented with two RTX 3090 GPUs. In stage II, Uformer\li B is served as backbone network and the learning rate of network is set as 0.0002. The Adam optimizer is adopted for optimization with a batch size of 48. The model was pre-trained for 200 epochs and fine-tuned for 100 epochs. The hyperparameters of image processing module are adjusted for each sequences individually.

\section{Experiment}\label{sec4}
\noindent\textbf{Dataset.} In stage I, only the provided 15 testing sequences are used to generate pseudo GT. In stage II, we use GT-Rain \cite{ba2022gt-rain} and WeatherStream-Rain \cite{zhang2023weatherstream} as a full version of training set. Specifically, 89 sequences from GT-Rain training set and 473 sequences from WeatherStream-Rain training set are combined together for pre-training. Then we fine-tune the final 15 sequences on pseudo GT generated in stage I. 

\subsection{Final Results on Testing Sequences}
Quantitatively, we rank 1st in terms of SSIM (0.8344) and 2nd in terms of PSNR	(26.604 dB), which reaches dominant position on the leaderboard. The visualization of deraining results from each stage of our pipeline are shown in Fig. \ref{Visualization}, which also demonstrates the superiority of our proposed two-stage deraining method. In stage I, comparing with multi frame average outcome, the pseudo GT generated by low-rank video deraining + image processing performs better at preserving image structure and removing haze-like veiling effect, especially the textured regions of case 1788\li 0 in the last row. Moreover, after processing by image processing module, pseudo GT could be generated with enhancing contrast and sharp edge. In stage II, the pre-trained Uformer already has strong generalization on testing sequences with few rain streaks and veiling effect left. Finally, the pre-trained model fine-tuned on previous pseudo GT further improve restoration with more visual pleasing background.  
 
\subsection{Ablation Study}
We conduct ablation study on our solutions to explore the contributions of each stage and module. The quantitative results of ablation study are listed in Table \ref{tab2}. Comparing with averaging, low-rank based deraining method performs better in term of multi frames deraining, and image processing module fully explores the potential of image restoration, improving 2.858 dB by adjusting image contrast and sharpness individually for each sequence. Moreover, though pre-trained model performs worse than pseudo GT for the remaining streak and veiling effect, the outcome increases up to 26.604dB/0.8344 after fine-tuning on the base of pseudo GT, which strongly supports that our solution is efficient and predominant in GT-Rain Challenge. 

 \begin{table}[htbp]
\normalsize
  \centering
  \caption{Ablation study on our proposed solution.}
    \setlength{\tabcolsep}{2.0mm}{
  \begin{tabular}{c|cc}
 \toprule
&PSNR&SSIM\\
 \midrule
Average&22.383&0.7779\\
Low-rank&23.158&0.7967\\
Pseudo GT&26.014&0.8283\\
Pre-trained&24.128&0.8056\\
\textbf{Fine-tuned}&\textbf{26.604}&\textbf{0.8344}\\
 \bottomrule
  \end{tabular}}
  \label{tab2}
 \end{table}
 
\section{Conclusion}
In our submission to the track 3 in UG$^{2}$+ Challenge in CVPR 2023, we propose an efficient two-stage image deraining framework as solution. In stage I, we adopt low-rank based video deraining method to generate high-quality pseudo GT without rain. In stage II, Uformer is utilized to pre-trained on GT-Rain and WeatherStream-Rain and then fine-tuned on pseudo GT to improve restoration. Finally, we rank 1st on SSIM and 2nd on PSNR. In the future, we will explore more efficient methods to improve this task.

{\small
\bibliographystyle{ieee_fullname}
\bibliography{egbib}

\begin{thebibliography}{1}\itemsep=-1pt

\bibitem{ba2022gt-rain}
Yunhao Ba, Howard Zhang, Ethan Yang, Akira Suzuki, Arnold Pfahnl,
  Chethan~Chinder Chandrappa, Celso de Melo, Suya You, Stefano Soatto, Alex
  Wong, and Achuta Kadambi.
\newblock Not just streaks: Towards ground truth for single image deraining.
\newblock In {\em ECCV}, 2022.

\bibitem{chang2017hyper}
Yi Chang, Luxin Yan, and Sheng Zhong.
\newblock Hyper-laplacian regularized unidirectional low-rank tensor recovery
  for multispectral image denoising.
\newblock In {\em CVPR}, pages 4260--4268, 2017.

\bibitem{fu2017removing}
Xueyang Fu, Jiabin Huang, Delu Zeng, Yue Huang, Xinghao Ding, and John Paisley.
\newblock Removing rain from single images via a deep detail network.
\newblock In {\em CVPR}, pages 3855--3863, 2017.

\bibitem{wang2019spatial}
Tianyu Wang, Xin Yang, Ke Xu, Shaozhe Chen, Qiang Zhang, and Rynson~WH Lau.
\newblock Spatial attentive single-image deraining with a high quality real
  rain dataset.
\newblock In {\em CVPR}, pages 12270--12279, 2019.

\bibitem{wang2022uformer}
Zhendong Wang, Xiaodong Cun, Jianmin Bao, Wengang Zhou, Jianzhuang Liu, and
  Houqiang Li.
\newblock Uformer: A general u-shaped transformer for image restoration.
\newblock In {\em CVPR}, pages 17683--17693, 2022.

\bibitem{yang2017deep}
Wenhan Yang, Robby~T Tan, Jiashi Feng, Jiaying Liu, Zongming Guo, and Shuicheng
  Yan.
\newblock Deep joint rain detection and removal from a single image.
\newblock In {\em CVPR}, pages 1357--1366, 2017.

\bibitem{zhang2023weatherstream}
Howard Zhang, Yunhao Ba, Ethan Yang, Varan Mehra, Blake Gella, Akira Suzuki,
  Arnold Pfahnl, Chethan~Chinder Chandrappa, Alex Wong, and Achuta Kadambi.
\newblock Weatherstream: Light transport automation of single image
  deweathering.
\newblock In {\em CVPR}, 2023.

\end{thebibliography}
}

\end{document}